\documentclass{article}

\PassOptionsToPackage{numbers,sort&compress}{natbib}
\usepackage[dblblindworkshop, final]{neurips_2026}

\usepackage[utf8]{inputenc}
\usepackage[T1]{fontenc}
\usepackage{hyperref}
\usepackage{url}
\usepackage{booktabs}
\usepackage{amsfonts}
\usepackage{amsmath}
\usepackage{amssymb}
\usepackage{nicefrac}
\usepackage{microtype}
\usepackage{xcolor}
\usepackage{graphicx}
\usepackage{tabularx}
\usepackage{algorithm}
\usepackage{algpseudocode}
\usepackage{enumitem}

\title{Temporal Expert Caching for Accelerating Inference in
MoE Diffusion Language Models}

\workshoptitle{DiffuLM: Diffusion Language Models --- Foundations, Efficiency, and Reasoning}

\author{%
  Vikhyath Kothamasu\\
  Carnegie Mellon University\\
  \texttt{vkothama@andrew.cmu.edu}\\
  \And
  Sandeep Kumar\\
  Carnegie Mellon University\\
  \texttt{sandeep3@andrew.cmu.edu}\\
  \And
  Deepika Palagani\\
  Carnegie Mellon University\\
  \texttt{dpalagan@andrew.cmu.edu}\\
}

\begin{document}
\maketitle

\begin{abstract}

Diffusion models have become the industry standard for image and video generation, with properties such as parallel decoding and bidirectional context distinguishing them from classical autoregressive alternatives. Diffusion language models (DLMs) bring the same principles to text, and recent models are competitive with autoregressive ones on code and mathematical reasoning. Sparse mixture-of-experts (MoE) variants such as the LLaDA series make these models more practical, offering large-model quality at a fraction of the per-token compute. On memory-constrained GPUs such as 12~GB consumer cards, however, the $\sim$32~GB of expert weights do not fit, and each of the dozens of denoising steps reloads them from host memory. Weight transfer, not compute, then dominates generation time. We identify two properties of iterative denoising that make most of this cost avoidable: between consecutive steps, per-token hidden states barely change (cosine similarity 0.97), and each layer routes to nearly the same set of experts (Jaccard overlap 0.94--0.97). We exploit this \emph{temporal locality} with three mechanisms: per-expert \emph{output caching} that skips unchanged tokens, \emph{temporal weight prefetching} driven by previous-step routing, and a CPU SwiGLU \emph{fallback} for misses. Across two LLaDA-mini models, two GPU architectures (L40S and A6000), and seven benchmarks, the combined system is a median \textbf{9$\times$} faster end-to-end than HF~Accelerate at a 12~GB budget, with task accuracy preserved within run-to-run variance.
\end{abstract}

\section{Introduction}
\label{sec:intro}

Diffusion language models (DLMs)~\cite{austin2021d3pm,sahoo2024mdlm,nie2025llada} generate text by iterative denoising rather than left-to-right decoding. Starting from a fully masked sequence, they iteratively refine all positions in parallel, attending to context in both directions. Recent models such as Dream~\cite{ye2025dream}, Mercury~\cite{inception2025mercury}, LLaDA2.0~\cite{bie2025llada20scalingdiffusionlanguage}, and LLaDA2.1~\cite{bie2026llada21speedingtextdiffusion} are competitive with autoregressive (AR) models on code, mathematical reasoning, and general knowledge. The LLaDA2 models further adopt a mixture-of-experts (MoE) design~\cite{shazeer2017outrageously,fedus2022switch} that makes them viable for real-world deployment: they reach 16~B parameters while activating only $\sim$1.4~B per token, since each token is routed to a handful of experts in every layer.

Parallel decoding with sparse compute is attractive for resource-constrained deployment, but that is also where these models are hardest to run. A single step uses only a few experts per layer, yet the full expert set is about 32~GB and does not fit in the memory of a consumer GPU. The rest of the experts live in host memory and must be streamed over PCIe, and this transfer, not compute, dominates inference time (Section~\ref{sec:motivation}). Masked diffusion makes the problem worse than in AR decoding. Rather than extending the sequence one token at a time, a DLM runs 32--64 denoising steps over the same sequence, each a full forward pass~\cite{bie2025llada20scalingdiffusionlanguage,bie2026llada21speedingtextdiffusion}. Inference systems built for AR generation, such as MoE-Infinity~\cite{xue2024moeinfinity}, FIDDLER~\cite{lu2025fiddler}, and HuggingFace Accelerate's \texttt{device\_map="auto"}~\cite{gugger2022accelerate}, treat each step independently. On memory-constrained GPUs, such as consumer cards with 12--24~GB of memory, they demand-load expert weights synchronously at every step and pay the full PCIe cost dozens of times per generation.

The same iterative structure, however, creates an opportunity that AR generation lacks. Consecutive denoising steps reprocess the same sequence with only small perturbations. Profiling LLaDA2.1-mini (Section~\ref{sec:motivation}), we find that per-token hidden states barely change between steps (cosine similarity 0.97) and that each layer routes to nearly the same experts (Jaccard overlap 0.94--0.97). Both the experts that fire and the outputs they produce are therefore largely redundant from one step to the next.

\paragraph{Contributions.}
We exploit this redundancy to skip a large majority of the repeated expert work, using three mechanisms:
\begin{itemize}[leftmargin=1.2em,itemsep=2pt,topsep=2pt]
\item \textbf{Per-expert output caching.} A cosine-similarity gate classifies each token into reuse, partial reuse, or recompute, and the dominant reuse class never loads expert weights.
\item \textbf{Temporal weight prefetching.} The prior step's routing is the prediction target, and transfers are staged across a pool of GPU buffers on independent streams, a configurable number of layers ahead of compute.
\item \textbf{CPU SwiGLU fallback.} Residual misses are served by sending $\sim$32~KB of activations to the host instead of pulling 6.3~MB of weights across PCIe.
\end{itemize}
% (iv) a fused Triton MoE kernel that consolidates per-expert dispatch into two launches per layer, reducing kernel launches by {up to 12$\times$}.

We evaluate on seven benchmarks from four task domains, two model versions (LLaDA2.0-mini and LLaDA2.1-mini), and two GPU architectures (L40S, Ada Lovelace; A6000, Ampere) at a fixed 12~GB GPU memory budget. The combined system is a median 9$\times$ faster than HF~Accelerate with task accuracy preserved within measured run-to-run variance, and the gains hold across both models, both GPUs, and all seven benchmarks.

\section{Related Work}
\label{sec:related}

\paragraph{MoE inference for autoregressive models.}
Most work on serving MoE models under a memory budget targets token-by-token decoding, and the central question is where expert weights live and when they move. MoE-Infinity~\cite{xue2024moeinfinity} traces activation patterns and uses them to decide which experts to keep cached and which to prefetch. FIDDLER~\cite{lu2025fiddler} takes a different route and runs mispredicted experts on the CPU instead of moving their weights. HOBBIT~\cite{tang2024hobbit} caches experts at several granularities and lowers precision when it has to evict. Pre-gated MoE~\cite{hwang2024pregated} separates expert selection from execution so that weight loads overlap with compute, while FineMoE~\cite{yu2025finemoe} and ExFlow~\cite{yao2024exflow} offload at finer granularity or follow inter-layer expert affinity. All of these treat every decoding step on its own, and since the step-to-step locality that diffusion inference provides does not exist in AR decoding, none of them can use it.

\paragraph{Reducing expert activations.}
TEAM~\cite{wei2026team}, the closest work to ours, observes the same temporal and spatial routing consistency in MoE DLMs but attacks a different bottleneck. With the whole model resident on a single 80~GB A100, TEAM activates fewer experts per forward pass to save computation, which changes routing and is therefore lossy. Our setting is memory-constrained, with a 12~GB budget that cannot hold the experts, so the cost is I/O rather than compute; we remove PCIe transfers and leave routing untouched. The foundational sparse-activation designs~\cite{shazeer2017outrageously,fedus2022switch} likewise control cost by limiting how many experts fire, whereas we keep the routed experts and only remove the cost of fetching them.

\paragraph{Diffusion-specific inference.}
Among the few systems that target diffusion language models directly, dLLM-Cache~\cite{liu2025dllmcacheacceleratingdiffusionlarge} reuses intermediate computations across denoising steps, but it is built for dense models and has no notion of per-expert reuse. dInfer~\cite{ma2025dinferefficientinferenceframework} is a multi-GPU framework aimed at decoding throughput rather than at fitting a model into limited memory. DiT-MoE~\cite{fei2024ditmoe} brings sparse experts to diffusion transformers for images, where sequences are far shorter. To our knowledge, no prior system exploits the step-to-step stability of hidden states and expert routing to serve MoE diffusion LMs efficiently under a memory constraint.

\section{Motivation}
\label{sec:motivation}

\paragraph{Breakdown of MoE-DLM inference time.}
% MoE-DLMs combine top-$k$ routing over a pool of 64--256 experts per layer (typically $k{=}8$) with 32--64 denoising passes over the same sequence: per-pass compute is sparse, yet is repeated dozens of times per generation. To locate the bottleneck in this workload, we profile a single forward pass of LLaDA2.1-mini at MoE-layer granularity using CUDA-event timing on H100. The expert-FFN component emerges as the dominant bottleneck, accounting for \textbf{87--89\%} of layer execution time -- set either by per-expert kernel launches when token counts are small, or by PCIe traffic when expert weights are streamed from host. Because masked diffusion repeats the full forward pass dozens of times over the same sequence, this expert-FFN bottleneck recurs at every denoising step. We next undertake an in-depth analysis of the structural properties of iterative denoising to identify how much of this repeated cost is recoverable.
An MoE-DLM routes each token to $k$ of the 64--256 experts in every layer (typically $k{=}8$), so a single forward pass is sparse. A full generation, however, consists of 32--64 such passes over the same sequence, and the expert work is repeated in each of them. To obtain a precise breakdown of this cost, we profiled a single forward pass of LLaDA2.1-mini at MoE-layer granularity using CUDA-event timing on an H100. We find that the expert FFNs account for \textbf{87--89\%} of layer execution time. When all expert weights fit in GPU memory, as on the H100, this time goes to per-expert kernel launches, since each expert sees only a few tokens. When the weights have to be streamed from host memory instead, PCIe traffic takes over. To measure the second case, we ran the HF~Accelerate baseline at a 12~GB budget on HumanEval+ with both models on both GPUs, and on GSM8K with both models on the A6000. Accelerate places each layer either on the GPU or in host memory, and an offloaded layer has its weights copied to the GPU every time it runs, so we timed the two kinds of layer separately. The offloaded layers take 94--96\% of per-step time, while the GPU-resident layers, which run the same computation with their weights already in place, take the remaining 4--6\%. The step time is therefore set by moving weights over PCIe rather than by running the experts, the same bottleneck that offloaded autoregressive MoE serving faces~\cite{xue2024moeinfinity,lu2025fiddler,hwang2024pregated}. Since every denoising step repeats the forward pass, this cost is paid dozens of times per generation, and our goal is to recover as much of it as possible. We next examine the structure of iterative denoising to see how much of that repeated work is redundant.

\paragraph{Profiling observations.}
% We instrument the 20 MoE layers of LLaDA2.1-mini (256 experts per layer, top-8 routing, BF16) with forward pre-hooks that record, at every denoising step, the layer-input hidden states and the top-$k$ expert indices emitted by each layer's gate. Generation uses gen-length 64 with 32 steps per block of 32 tokens, yielding 64 total denoising steps over the same sequence. The three observations below are averaged over all MoE layers and consecutive step pairs; formal definitions of every quantity reported are given in Appendix~\ref{app:profiling}.
For this analysis, we instrument the 20 MoE layers of LLaDA2.1-mini (256 experts, top-8, BF16) with forward pre-hooks that record the hidden states entering each layer and the top-$k$ expert indices chosen by its gate at every denoising step. Unless noted otherwise, the numbers below come from GSM8K prompts generated with the same settings as the main evaluation in Section~\ref{sec:setup} (32 denoising steps, generation length 384), averaged over all layers and all pairs of consecutive steps. Appendix~\ref{app:profiling} gives the formal definitions and the values for a shorter generation length.
\begin{enumerate}[leftmargin=1.2em,itemsep=2pt,topsep=2pt]
    \item \textbf{Routing stability.} We hypothesize that the set of experts a layer uses changes little from one step to the next, and we test this at three granularities because each bears on a different mechanism. At the \emph{layer} level, the union of experts chosen by any token, which is what must be resident on the device and hence what a prefetcher has to predict, mean Jaccard overlap between adjacent steps is \textbf{0.94--0.97} across the configurations we tried and \textbf{0.97} in the evaluation configuration. At the \emph{per-token} level, a token's own top-$k$ set compared with its set at the previous step, which decides whether that token's expert outputs can be reused, the overlap is \textbf{0.87--0.91}. The strictest test is the \emph{active block}. LLaDA generates in blocks of 32 tokens, and only the current block is being denoised at any step, so its tokens change the fastest. Restricted to these tokens, the overlap drops to \textbf{0.76--0.78} (0.70--0.82 across layers), which we take as the conservative bound. \emph{Under every definition, the previous step's routing predicts most of the next step's routing, and the prediction is free, since the gate has already produced those indices.}
    \item \textbf{State similarity.} For a given token and layer, the cosine similarity between the layer's input hidden state at one step and at the next averages \textbf{0.97}, and \textbf{85\%} of token-steps reach at least 0.97, including tokens that are still being denoised. \emph{Because the inputs barely move, the per-expert outputs barely move either, and they can be reused instead of recomputed.}
    % REVIEW NOTE: ~78 experts/layer implies ~10 GB of resident expert weights; Section 4.2 says N=5 buffers of 1.6 GB (~8 GB, ~63 experts/layer). Reconcile the two.
    \item \textbf{Expert sparsity and working-set size.} Token assignments are highly uneven at every layer and step, with the top 10 experts per layer receiving about \textbf{65\%} of them. The active set is nonetheless large in absolute terms, since at generation length 384 an average of \textbf{$\sim$150 of 256} experts per layer receive at least one token per step. \emph{Routing mass is concentrated and, by Observation~1, stable over time. But a 12~GB budget can keep only about 78 experts per layer resident once the rest of the model is placed, so the working set does not fit, residency alone cannot avoid transfers, and the expert set has to be predicted.}
\end{enumerate}

\section{Methodology}
\label{sec:method}

Based on the redundancies identified in Section~\ref{sec:motivation}, we propose two techniques: \textbf{per-expert output caching} to exploit state similarity, and \textbf{temporal expert weight prefetching} with a CPU-resident SwiGLU fallback to exploit routing stability and sparsity. A combined architecture runs these both within a single forward pass such that each expert evaluation is served from the output cache (Tier~1), from a prefetched GPU weight buffer (Tier~2), or by the CPU fallback (Tier~3).

\subsection{Per-Expert Output Caching}
\label{sec:method:cache}

A lightweight per-layer cache ($\sim$40MB) stores the outputs, routing IDs, and input hidden states of top-$k$ experts. At step $t{+}1$, tokens are categorized using cosine-similarity threshold $\tau$ against step $t$:
\begin{itemize}[leftmargin=1.2em,itemsep=2pt,topsep=2pt]
\item \textbf{Cat A (Full Reuse)}: Same experts, similar input ($> \tau$). Cached outputs are re-weighted by current gate; \emph{no expert work.}
\item \textbf{Cat B (Partial Reuse)}: Similar input, routing partially changed. Reuses shared experts and recomputes only newly-routed ones.
\item \textbf{Cat C (Full Recompute)}: Significant input change. Standard MoE forward pass.
\end{itemize}

At $\tau{=}0.99$ with naive reuse, the cache eliminates expert work for 64--88\% of tokens (Section~\ref{sec:results}), removing both PCIe overhead and compute.

\subsection{Temporal Expert Weight Prefetching}
\label{sec:method:prefetch}

Expert weights ($\sim$32~GB for LLaDA2.1-mini) exceed the 12--24~GB consumer GPU budget and reside in pinned host memory. Individual experts are copied to device over PCIe on demand. 

Let $E_\ell^t \subseteq \{1,\ldots,256\}$ denote the set of experts selected for at least one token at MoE layer~$\ell$ and step~$t$ (typically $|E_\ell^t|\!\approx\!150$ at the operating point, Section~\ref{sec:motivation}). Every uncached expert in $E_\ell^{t+1}$ places its PCIe transfer on the compute critical path, stalling layer~$\ell$'s computations. 

We instead issue transfers on independent CUDA streams ahead of time, using $E_\ell^t$ to predict $E_\ell^{t+1}$. This works because the prefetch target is exactly the layer-level active set, whose Jaccard overlap between consecutive steps is $0.94$--$0.97$ (Section~\ref{sec:motivation}). Appendix~\ref{app:algorithm} gives the resulting three-tier pipeline as a schematic (Figure~\ref{fig:prefetch}) and pseudocode.

\paragraph{Temporal predictor.}
We take $\hat E_\ell^{t+1} \stackrel{\text{def}}{=} E_\ell^t$: experts routed at layer~$\ell$ during step~$t$ become the prefetch target for the same layer at step~$t{+}1$. The predictor has no learned parameters and reuses only the previous step's routing decisions. Experts in $E_\ell^{t+1}\setminus\hat E_\ell^{t+1}$ are residual misses handled by the fallback below.

% \paragraph{Multi-stream buffer pool with layer-level lookahead.}
% The prefetch path maintains $N$ on-device buffers, each holding up to $C{=}256$ experts ($C\cdot W_e\!\approx\!1.6$~GB, with $W_e\!\approx\!6.3$~MB per expert). $N$ is set by however much GPU memory remains after accounting for non-expert state, giving $N{=}5$ ($\approx$8~GB) at a 12~GB budget. A single \emph{compute stream}~$\sigma_c$ runs attention, gating, the fused MoE matmuls, and unsort. Alongside it, $N$ \emph{transfer streams} are bound one-to-one to buffers, letting the DMA engine interleave several layers' host-to-device copies instead of serializing them behind $\sigma_c$. A lookahead parameter $D$ controls how many MoE layers are staged ahead of compute. Concretely, the scheduler enqueues the copies for $\hat E_\ell^{t+1}$ on the assigned buffer's stream and records a CUDA event. When $\sigma_c$ reaches layer~$\ell$, it waits on that event before dispatching matmuls, a GPU-side dependency rather than a host synchronization. Completed buffers are then recycled to layer~$\ell{+}D$. If $D\!>\!N$, the scheduler preempts the buffer with the largest scheduling slack, and the preempted transfer is either reissued or served via the fallback. The matmul itself uses a fused single-launch SwiGLU Triton kernel, which eliminates the per-expert launch overhead that would otherwise dominate in the small-token regime. The per-step PCIe volume is $\sum_\ell |\hat E_\ell^{t+1}|\cdot W_e$, determined entirely by the temporal predictor.

\paragraph{Multi-stream buffer pool with layer-level lookahead.}
Prefetched weights land in a pool of $N$ GPU buffers, each buffer holds up to $C{=}256$ experts (1.6~GB at $W_e\!\approx\!6.3$~MB per expert). $N$ is set by the GPU memory left after non-expert weights: $N{=}5$ ($\approx$8~GB) at a 12~GB budget. Work runs on two kinds of CUDA streams: a single \emph{compute stream}~$\sigma_c$ handles attention, gating, and the MoE matmuls, while $N$ \emph{transfer streams}, one per buffer, copy weights from host memory. Because each buffer has its own stream, copies for several upcoming layers can run at the same time instead of queuing behind compute. The lookahead $D$ sets how many layers ahead of compute these copies are staged. For each upcoming layer~$\ell$, the scheduler queues the copies for $\hat E_\ell^{t+1}$ on that layer's buffer stream and records a CUDA event. When $\sigma_c$ reaches layer~$\ell$, it waits on that event on the GPU, without stalling the host. Once layer~$\ell$ finishes, its buffer is reused for layer~$\ell{+}D$. If $D\!>\!N$, there are more layers to stage than buffers, so the scheduler takes over the buffer whose layer is furthest away, and that layer's transfer is either reissued or served by the fallback. The expert matmuls run in a single fused SwiGLU Triton kernel, which avoids the per-expert launch overhead that would otherwise dominate when each expert sees only a few tokens. The PCIe volume per step is therefore $\sum_\ell |\hat E_\ell^{t+1}|\cdot W_e$, set entirely by the temporal predictor.

\paragraph{CPU SwiGLU fallback.}
A residual miss at layer~$\ell$ is an expert $e \in E_\ell^{t+1}\setminus\hat E_\ell^{t+1}$; synchronously fetching its 6.3~MB weight would add $\sim$0.35~ms to the critical path. The fallback inverts the transfer direction: the activation tile for $e$ ($|\text{tokens}_e|\!\cdot\!H\!\cdot\!2$~B, $\approx$32~KB at typical token counts) is copied to host, SwiGLU runs on the pinned host weights via a CPU thread pool, and outputs are copied back. Because per-miss volume is $\sim$200$\times$ smaller than the corresponding weight transfer, the round-trip is issued on a transfer stream and overlaps with the remainder of $\sigma_c$.

\subsection{Combined Three-Tier System}
\label{sec:method:combined}

The output cache (Section~\ref{sec:method:cache}) and temporal weight prefetching (Section~\ref{sec:method:prefetch}) are complementary: the cache eliminates both computation and weight access for Cat~A tokens, while prefetching reduces expert-weight latency for Cat~B and Cat~C tokens.

\begin{figure*}[t]
  \centering
  \includegraphics[width=0.8\textwidth]{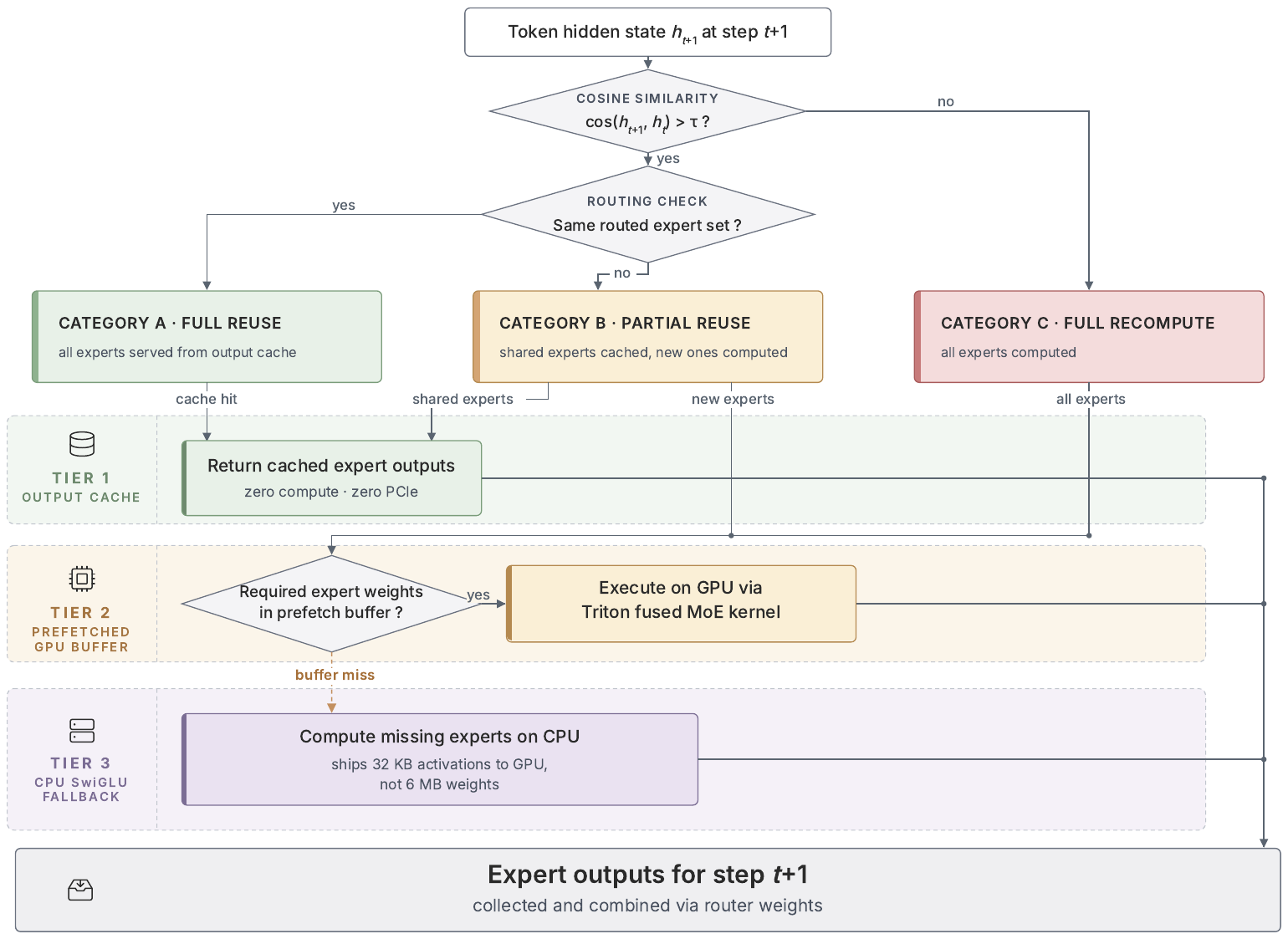}
  \caption{Combined three-tier dispatch at step $t{+}1$. Each token is classified using hidden-state cosine similarity and routing-set overlap, producing Cat~A (full cache reuse), Cat~B (partial reuse), or Cat~C (full recompute). Cat~B/C computation first attempts execution from the prefetched GPU expert buffer via the fused Triton MoE kernel, while residual misses are handled through the CPU SwiGLU fallback.}
  \label{fig:combined}
\end{figure*}

We integrate both mechanisms into a unified forward pass, summarized in Figure~\ref{fig:combined}. For each MoE layer at denoising step $t{+}1$:

\begin{itemize}[leftmargin=1.2em,itemsep=2pt,topsep=2pt]
\item \textbf{Cat~A (Full Reuse):} Tokens are served entirely from cache; expert weights are never touched.
\item \textbf{Cat~B (Partial Reuse):} Shared experts reuse cached outputs, while newly-routed experts are served from the prefetched GPU buffer, falling back to CPU SwiGLU on miss.
\item \textbf{Cat~C (Full Recompute):} All experts are executed through the prefetched GPU path or CPU fallback.
\end{itemize}

The cache also reshapes the prefetching problem itself: Cat~A experts become invisible to the temporal predictor because they no longer require expert execution. As a result, the prediction set shrinks, allowing the prefetcher to concentrate its limited on-device buffer capacity on the subset of experts that still require transfers, which reduces aggregate PCIe traffic.

\section{Experimental Setup}
\label{sec:setup}

\subsection{Models}
We evaluate two MoE DLMs from the LLaDA family, LLaDA2.0-mini~\cite{bie2025llada20scalingdiffusionlanguage} and LLaDA2.1-mini~\cite{bie2026llada21speedingtextdiffusion}, to validate that our system is model-version-agnostic. Both share the same sparse-MoE architecture (256 experts per MoE layer, top-8 routing, 20 layers, hidden 2048, intermediate 512, BF16), totaling $\sim$16~B parameters with $\sim$1.4~B active per token.

\subsection{Hardware}
We evaluate on two distinct GPU architectures to test hardware agnosticism: \textbf{L40S} (Ada Lovelace, 48~GB GDDR6, 864~GB/s bandwidth, 362 dense BF16 tensor~TFLOPS) and \textbf{A6000} (Ampere, 48~GB GDDR6, 768~GB/s bandwidth, 155 dense BF16 tensor~TFLOPS). To simulate a low-resource consumer-GPU, we cap usable VRAM at 12~GB (enforced via Accelerate's per-device memory cap), forcing a 63\% offload ratio. Host RAM ($\sim$120~GB) is sufficient for pinned expert weights. 

\subsection{Benchmarks}
\label{sec:setup:benchmarks}
We span four task domains chosen to stress generalization. Prompting protocols, generation lengths, and scoring rules are in Appendix~\ref{app:benchmarks}.

\begin{itemize}[leftmargin=1.2em,itemsep=2pt,topsep=2pt]
\item \textbf{Code Generation}: HumanEval+~\cite{liu2023humanevalplus}, $n{=}164$; LiveCodeBench easy+medium~\cite{jain2024lcb}, $n{=}310$; both pass@1
\item \textbf{General Knowledge}: ARC-C~\cite{clark2018arc}, $n{=}1{,}172$; MMLU~\cite{hendrycks2021mmlu}, $n{=}14{,}042$ across 57 subjects; TriviaQA~\cite{joshi2017triviaqa}, $n{=}1{,}000$
\item \textbf{Mathematical Reasoning}: GSM8K~\cite{cobbe2021gsm8k}, $n{=}1{,}319$ 
\item \textbf{Long Compositional Reasoning}: BBH~\cite{suzgun2022bbh}, $n{=}6{,}511$ across 23 subtasks. 
\end{itemize}

\subsection{Methods Compared}
\label{sec:setup:methods}
On every benchmark, we compare two end-to-end serving strategies at fixed 12~GB GPU memory:

\paragraph{HF Accelerate baseline.}
HF Accelerate with automatic device placement demand-loads non-resident expert weights synchronously over PCIe at every denoising step. This is the default for memory-constrained MoE serving. Standard MoE inference frameworks (e.g.\ MoE-Infinity, FIDDLER, HOBBIT) are tuned for autoregressive serving and do not translate to the multi-step, single-sequence structure of masked-diffusion inference, so HF~Accelerate is used as the  drop-in baseline for this regime.

\paragraph{Combined three-tier system.}
Our full system (Section~\ref{sec:method:combined}): per-expert output caching, temporal weight prefetching with the CPU SwiGLU fallback, and the fused Triton MoE kernel, integrated into a single forward pass. End-to-end speedup is the ratio of Accelerate to combined per-problem wall time. Table~\ref{tab:three-mode} separately reports per-step latency ratios.

\paragraph{Intermediate prefetch mode (HumanEval+, GSM8K).}
On these two benchmarks we additionally report the prefetch configuration (no output cache), which includes temporal weight prefetching, the CPU SwiGLU fallback, and the fused Triton MoE kernel. These comparisons show the speedup progression over HF~Accelerate: accel~$\ll$~prefetch~$<$~combined (Table~\ref{tab:three-mode}). We observe the same speedup ordering across the other evaluated benchmarks.

\subsection{Preliminary Ablations}
\label{sec:setup:ablations}
Three hyperparameters, the cosine-similarity threshold $\tau$, the prefetch lookahead $D$, and the routing predictor, were fixed by ablations prior to the main evaluation (Appendix~\ref{app:ablations}). 

\begin{itemize}[leftmargin=1.2em,itemsep=2pt,topsep=2pt]
\item First, naive reuse at $\tau{=}0.99$ gives the largest speedup at no accuracy loss (tighter thresholds and the Jacobian variant lose net throughput)
\item Second, $D{=}3$ minimizes p50 ms/step at every (model, GPU-budget) cell, with deeper lookahead adding 19--36\% latency without raising hit rate.
\item Third, the pure-temporal predictor outperforms spatial, deep, and EMA-workload augmentations, which raise hit rate by 1--4~pp but slow ms/step by 1--64\%. 
\end{itemize}

\subsection{Configuration}
Generation runs at temperature 0 with top-$k{=}8$ routing, 32 diffusion steps, and block length 32. Following the ablations of Section~\ref{sec:setup:ablations}, the output cache uses naive reuse at $\tau{=}0.99$, the prefetcher uses lookahead $D{=}3$ over a 5-buffer pool, and routing predictions come from previous-step routing. Code execution for coding tasks runs in a sandboxed subprocess with a 60s timeout per problem.

\section{Results}
\label{sec:results}

We report end-to-end speedup of the combined system compared to the accelerate baseline for all benchmarks, models, and GPUs, together with the quality metric of each benchmark. For HumanEval+ and GSM8K, we also report per-step latency for the intermediate prefetch configuration without output caching.

\subsection{End-to-End Speedup}
\label{sec:results:speedup}

Across the four (model, GPU) combinations and seven benchmarks, the combined system runs 5.0--14.4$\times$ faster end to end than HF~Accelerate, with a median of \textbf{9.2$\times$} and an arithmetic mean of 9.3$\times$ (Table~\ref{tab:headline}). We compute each speedup as Accelerate's per-problem wall time divided by the combined system's, and all 28 comparisons receive equal weight in these summaries. Table~\ref{tab:three-mode} breaks out per-step latency for Accelerate, prefetch, and combined on the L40S, using HumanEval+ and GSM8K. Prefetch alone, which includes the CPU SwiGLU fallback and the fused Triton kernel, is 4.0--6.7$\times$ faster than Accelerate. Adding output caching pushes that to 8.1--10.9$\times$, demonstrating its additional benefit over prefetch.

\begin{table}[t]
\centering
\footnotesize
\setlength{\tabcolsep}{5pt}
\caption{Per-step latency on L40S at 12~GB. For each model, the top row reports ms/step and the bottom row reports speedup over HF~Accelerate.}
\label{tab:three-mode}
\begin{tabular}{llccc}
\toprule
\textbf{Benchmark} & \textbf{Model} & \textbf{accel} & \textbf{prefetch} & \textbf{combined} \\
\midrule
HumanEval+   & 2.0-mini & 5086        & 758         & 612          \\
      &          & 1.0$\times$ & 6.7$\times$ & 8.3$\times$  \\
\cmidrule(l){2-5}
      & 2.1-mini & 4764        & 751         & 585          \\
      &          & 1.0$\times$ & 6.3$\times$ & 8.1$\times$  \\
\midrule
GSM8K & 2.0-mini & 3931        & 790         & 362          \\
      &          & 1.0$\times$ & 5.0$\times$ & 10.9$\times$ \\
\cmidrule(l){2-5}
      & 2.1-mini & 2959        & 735         & 363          \\
      &          & 1.0$\times$ & 4.0$\times$ & 8.2$\times$  \\
\bottomrule
\end{tabular}
\end{table}

\begin{table}[t]
\centering
\small
\setlength{\tabcolsep}{4pt}
\caption{End-to-end speedup over HF~Accelerate at the 12~GB memory budget across seven benchmarks and four (model, GPU) combinations.}
\label{tab:headline}
\begin{tabular}{lcccc}
\toprule
 & \multicolumn{2}{c}{\textbf{LLaDA2.0-mini}} & \multicolumn{2}{c}{\textbf{LLaDA2.1-mini}} \\
\cmidrule(lr){2-3}\cmidrule(lr){4-5}
\textbf{Benchmark} & \textbf{L40S} & \textbf{A6000} & \textbf{L40S} & \textbf{A6000} \\
\midrule
HumanEval+    & 8.3$\times$  & 12.4$\times$ & 8.1$\times$  & 5.6$\times$  \\
LiveCodeBench & 11.5$\times$ & 11.3$\times$ & 9.2$\times$  & 12.5$\times$ \\
GSM8K         & 10.9$\times$ & 12.4$\times$ & 8.2$\times$  & 11.3$\times$ \\
ARC-C         & 5.5$\times$  & 9.8$\times$  & 5.6$\times$  & 8.4$\times$  \\
MMLU          & 9.7$\times$  & 10.3$\times$ & 5.6$\times$  & 5.2$\times$  \\
TriviaQA      & 7.0$\times$  & 14.4$\times$ & 14.3$\times$ & 8.6$\times$  \\
BBH           & 14.2$\times$ & 5.8$\times$  & 5.0$\times$  & 9.1$\times$  \\
\bottomrule
\end{tabular}
\end{table}

\paragraph{Quality preservation.}
Temporal prefetching is \emph{lossless}, since it gives exactly the same results as the baseline, as it only changes where the expert weights live and the routed experts are still computed exactly. The output cache is an \emph{approximation}, because Cat~A tokens reuse the expert outputs from a previous step and re-weight them with the current gate. We decode at temperature 0, so these differences do not come from sampling, and we measured them directly. To put the size of the drift in context, we first checked how much the system varies against itself. We ran an identical configuration (LLaDA2.1-mini, L40S, 12~GB) three times and got 85.5/85.5/86.5\% on a 200-problem GSM8K subset and 80.5/83.5/82.3\% on HumanEval+. Between any two of these runs, 8--10\% of examples flip correctness, and they flip in both directions about equally often. We attribute this to floating-point reduction order in the multi-stream runtime, which threshold-based adaptive stepping amplifies until the model follows a different but equally valid trajectory. Measured against that run-to-run spread, combined caching stays close to the accelerate baseline. On the coding benchmarks, the performance gap is at most 3.5~percentage points. On GSM8K, the exact-match accuracy gap is 0.3~percentage points for LLaDA2.0-mini (90.0\% vs.\ 90.3\%) and 1.3~percentage points for LLaDA2.1-mini (85.6\% vs.\ 84.3\%). We also checked this on the full test set, covering all 1{,}319 GSM8K problems, using LLaDA2.0-mini on the A6000, where the baseline scores 89.69\% (95\% CI [88.05, 91.33]) and combined caching scores 90.45\%, which falls within the baseline interval. The full per-cell table is in Appendix~\ref{app:accuracy}.

\subsection{Cache Mechanism Across Benchmarks}
\label{sec:results:cache}

The cache-hit fraction (Cat~A) sets an upper bound on how much expert work the cache can remove, and it ranges from 64\% on HumanEval+ to 88\% on MMLU/GSM8K (Appendix~\ref{app:cat-table}, Table~\ref{tab:cat}; Figure~\ref{fig:cat}). On every benchmark, a large share of expert recomputation can be skipped whenever consecutive denoising steps see nearly identical inputs. Cat~B never exceeds 2.1\% on any benchmark. At $\tau{=}0.99$, a token whose routing changes has almost always drifted far enough in input space to fail the cosine threshold as well, so it falls through to Cat~C, which absorbs the remainder.

\subsection{GPU Agnosticism}
\label{sec:results:gpu}

Per-cell speedups shift with GPU throughput and PCIe bandwidth, but both GPUs give a large gain in every cell. For a fixed model and benchmark, the Cat~A fraction on the L40S and the A6000 agree to within 0.5~pp. This suggests the cache works because of how the diffusion process behaves and has little to do with the hardware it runs on.

\section{Analysis}
\label{sec:analysis}

\paragraph{The system is PCIe-bandwidth-bound, not latency-bound.}
The lookahead sweep (Appendix~\ref{app:abl:lookahead}, Table~\ref{tab:abl:lookahead}) shows that increasing $D$ from 3 to 12 raises wall time by 19--36\%, with $D{=}3$ achieving the lowest wall time in every configuration. At $D{=}3$, expert transfers already bring the PCIe link close to saturation, leaving little bandwidth headroom for deeper lookahead to exploit. Under the 12~GB budget, these transfers total about 18.7~GB per denoising step ($\approx$150 experts per layer $\times$ 20 layers $\times$ 6.3~MB; Section~\ref{sec:motivation}), corresponding to a sustained throughput of roughly 24~GB/s, which is near the PCIe Gen4 ceiling on both the L40S and A6000. A larger $D$ queues more DMA transfers without getting them through the link any faster, and once $D$ exceeds $N{=}5$, it also starts stealing buffers. We see the same thing at the 24~GB budget, where the hit rate for LLaDA2.1-mini stays exactly flat across $D \in \{3,6,12\}$ because the bigger buffer pool has nothing extra to retire. That tells us $D{=}3$ already covers everything the link can carry in flight. The speedup comes from transferring about 75\% less data over PCIe per step, not from hiding transfer latency. As a sanity check, $18.7~\text{GB} / 24~\text{GB/s} \approx 780$~ms matches the 735--790~ms/step \emph{prefetch} column of Table~\ref{tab:three-mode} within $\sim$6\% across all four (benchmark, model) cells. The \emph{combined} column (362--612~ms) does not follow this curve. Output-cache reuse removes about 88\% of the per-step transfers because Cat~A bypasses PCIe entirely, so per-step volume falls to roughly 2.3~GB and the bottleneck shifts to compute and the non-MoE layers.

\paragraph{Why pure temporal prediction is the default.}
We compared three augmented predictors with pure temporal prediction (Appendix~\ref{app:abl:predictor}, Table~\ref{tab:abl:predictor}). The \emph{temporal-spatial} variant also uses the routing in the previous step of the adjacent layers ($\ell{\pm}1$), and it reduces the miss rate in step~0 from 16.2\% to 8.4\%, which halves the spike at the block boundaries. The \emph{EMA-workload-expanded} variant adds the top-$K$ experts ranked by an exponentially weighted moving average of per-step token counts. That lowers the steady-state median miss rate from 10.9\% to 6.9\%, a 30\% reduction. The \emph{temporal-deep} variant adds the same layer's routing from two steps back, and it barely helps either the cold start or the steady state. The boundary spike occurs when 32 fresh tokens enter the active set with no routing history, and none of the three variants eliminates it. Beyond failing to remove that spike, all three variants make each denoising step slower, adding between 1 and 64\% to the ms/step. We kept pure temporal as the default, because in the PCIe-bound regime described above a lower miss rate on its own does not reduce end-to-end latency. Combining temporal-spatial with EMA-expanded looks like a natural next step, but we leave it to future work.

\paragraph{Cat~A reuse depends on output style, not generation length alone.}
Cat~A ranges from 64\% on HumanEval+ to 88\% on MMLU across the benchmarks (Appendix~\ref{app:cat-table}, Table~\ref{tab:cat}). Generation length does not explain that spread on its own, since HumanEval+ and BBH both generate at length 512 and still end up at 64\% and 86\%. The real difference lies in the style of the output, because free-form chain-of-thought and prose (GSM8K, MMLU, BBH) change smoothly from one denoising step to the next, and so do the long LiveCodeBench solutions, and all of these reach 84--88\%. Structured code completion and short discrete answers (HumanEval+, TriviaQA, ARC-C) commit their tokens in larger jumps, which is why they land at only 64--68\%. The output cache therefore benefits the most on long-form natural-language generation.

\paragraph{Cat~B (Jacobian correction) is unnecessary at $\tau{=}0.99$.}
Cat~B covers tokens whose input is similar but whose routing has changed, and it accounts for $\leq$2.1\% of tokens across all seven benchmarks. This is because a token whose routing changes has nearly always drifted in input space as well, so it ends up in Cat~C instead. Given how little Cat~B holds, we disable the Jacobian correction path by default. The $\tau$ sweep (Appendix~\ref{app:abl:tau}, Table~\ref{tab:abl:tau}) supports our choice of operating point, because tightening to $\tau{=}0.999$ shrinks Cat~A by about 6~pp without any gain in accuracy. The Jacobian variant is even worse, since it runs at 0.49$\times$ the speed of the uncached baseline and so ends up being slower than having no cache at all.

\section{Conclusion}
\label{sec:conclusion}
% We presented a system for accelerating inference in MoE diffusion language models that exploits two structural properties unique to iterative denoising: cross-step routing stability and high cosine similarity of intermediate states. Across two model versions, two GPU architectures, and two distinct task domains (code generation and reasoning), our combined caching delivers a \textbf{median 9--12$\times$ end-to-end speedup} over HF~Accelerate at matched 12~GB memory budgets, with output quality preserved within sampling noise.

We built a system that speeds up inference for MoE diffusion language models by exploiting two properties specific to iterative denoising, namely that routing stays stable across steps and that intermediate states have high cosine similarity. With a 12~GB budget, combined caching gives a median \textbf{9$\times$} end-to-end speedup over HF~Accelerate, and task accuracy stays within the run-to-run variance we measured. Along the way, we reached three findings that we expect to hold beyond this particular setup. (i) In this PCIe-bound regime, the number of bytes moved per step matters most, while latency matters much less. (ii) The remaining misses are structural, because they cluster at block boundaries where routing shifts wholesale, which is why no augmentation based on routing history can anticipate them. (iii) Reuse depends on output style more than on generation length alone, since free-form chain-of-thought benchmarks reach 84--88\% Cat~A while structured code completion reaches only 64\% at the same generation length. Because of this, the approach pays off most when the task is long-form natural-language generation rather than short structured output.

% Pure temporal prediction sits on the Pareto frontier; no history-based augmentation beats it, because residual misses concentrate at block boundaries where routing changes wholesale
% the predictor's residual misses are structural, concentrating at block boundaries where routing shifts wholesale, so no history-based augmentation can anticipate them---the augmentations we tested each raised hit rate by only 1--4~pp while enlarging the prefetch set;

% Three findings generalize beyond our system: (i) in this PCIe-bound regime the dominant lever is bytes transferred per step---optimizations that did not reduce PCIe volume yielded no end-to-end gain; (ii) pure temporal prediction sits on the Pareto frontier, since residual misses concentrate at block boundaries where routing changes wholesale and no history-based augmentation anticipates them; and (iii) the caching benefit grows with generation length (\pending{20~pp} higher Cat~A at gen=1024 than gen=512), making the approach increasingly valuable for longer-context tasks.
\section*{Limitations}

% \paragraph{Single-GPU, single-sequence scope.}
% The system has not been integrated with batched or distributed serving frameworks (e.g., dInfer~\cite{ma2025dinferefficientinferenceframework} with tensor parallelism). Scaling temporal prefetching to high-throughput batched serving requires managing concurrent routing predictions and per-sequence cache pools, an open systems problem we leave to future work.

% \paragraph{Bounded utility on unconstrained hardware.}
% Our techniques target memory-constrained scenarios where expert weights must be paged across PCIe. On GPUs that hold the full MoE in VRAM the prefetch and fallback paths become inert, and only the per-expert output cache contributes a residual saving by skipping the recomputation of Cat-A experts.

% \paragraph{Two-domain evaluation.}
% We evaluate on two domains (coding, reasoning) and three benchmarks. Extending to retrieval-style tasks, conversational generation, and tool-use settings is the natural next step.

Everything we measured ran on a single GPU serving one sequence at a time. This is the situation that a memory-constrained user is most likely to face, but it tells us little about how the system would behave in production. Extending temporal prefetching to batched or distributed serving (for instance, dInfer~\cite{ma2025dinferefficientinferenceframework} with tensor parallelism) would require tracking routing predictions for many sequences at once and maintaining a separate cache pool for each. We leave this extension to future work. The speedups also depend on memory pressure: if the GPU can hold the entire MoE in VRAM, the prefetch and fallback paths never run and only the output cache helps. Our model coverage is narrower than our benchmark coverage because the seven benchmarks span four kinds of tasks while both models (LLaDA2.0-mini and LLaDA2.1-mini) share a single architecture. We do not yet know whether the results hold for a structurally different MoE-DLM such as SDAR~\cite{cheng2025sdar} or for retrieval, conversational, and tool-use workloads, and we leave both questions to future work.

\bibliographystyle{plainnat}
\bibliography{custom}

\newpage
\appendix

\section{Profiling Definitions}
\label{app:profiling}

This appendix gives precise definitions for the three quantities measured in Section~\ref{sec:motivation}.

\subsection{Notation}
\label{app:notation}

We adopt the following notation throughout, instantiated for LLaDA2.1-mini in the profiling setup of Section~\ref{sec:motivation}:

\begin{itemize}\setlength\itemsep{2pt}
\item $L$ --- number of MoE layers ($L{=}20$).
\item $E$ --- number of experts per MoE layer ($E{=}256$).
\item $k$ --- routing fan-out (top-$k$ experts per token, $k{=}8$).
\item $T$ --- number of denoising steps per generation ($T{=}64$ in the profiling setup; the main experiments of Section~\ref{sec:setup} use $T{=}32$, the model's recommended configuration).
\item $S$ --- sequence length (prompt plus generation tokens).
\item $d$ --- model hidden dimension ($d{=}2048$).
\item $\ell \in \{1,\ldots,L\}$ --- MoE layer index.
\item $t \in \{0,\ldots,T-1\}$ --- denoising step index.
\item $H_\ell^t \in \mathbb{R}^{S \times d}$ --- input hidden states to MoE layer $\ell$ at step $t$, recorded by a forward pre-hook on the MoE block.
\item $R_\ell^t \in \{1,\ldots,E\}^{S \times k}$ --- top-$k$ expert indices selected by the gate of layer $\ell$ at step $t$ for each token.
\item $E_\ell^t = \bigl\{ R_\ell^t[s, j] : s \in [S],\, j \in [k] \bigr\} \subseteq \{1,\ldots,E\}$ --- the set of \emph{active} experts at layer $\ell$, step $t$ (experts selected for at least one token).
\end{itemize}

\subsection{Routing Stability (Observation~1)}
\label{app:routing}

The Jaccard overlap between active expert sets at consecutive steps is
\begin{equation*}
J_\ell^t \;=\; \frac{\bigl|E_\ell^t \cap E_\ell^{t+1}\bigr|}{\bigl|E_\ell^t \cup E_\ell^{t+1}\bigr|} \in [0, 1].
\end{equation*}
We evaluate this at three granularities, differing only in which token set defines $E_\ell^t$:
\begin{itemize}\setlength\itemsep{2pt}
\item \textbf{Layer-level}: $E_\ell^t$ is the union over all $S$ sequence positions, as defined above. This is the set that must be resident on device, so it is the quantity the prefetcher targets.
\item \textbf{Per-token}: $J$ is computed separately for each token's own top-$k$ set, $\{R_\ell^t[s,:]\}$ against $\{R_\ell^{t+1}[s,:]\}$, then averaged over $s$. This is the quantity that governs whether a given token's expert outputs can be reused.
\item \textbf{Active-block}: $E_\ell^t$ is the union restricted to the 32 positions currently being denoised. These tokens change fastest, so this is the conservative bound.
\end{itemize}

\noindent Measured on GSM8K prompts with LLaDA2.1-mini, means are 0.94--0.97 (layer-level), 0.87--0.91 (per-token), and 0.76--0.78 (active-block, 0.70--0.82 across layers). Each moves by at most 0.016 between $T{=}64$ and $T{=}32$, so the premise does not depend on the number of denoising steps, and all three are strongest at the benchmark operating point (32 steps, generation length 384). We report all three because the paper's headline claims rest on different ones: layer-level for weight prefetching, per-token for output reuse.

\subsection{State Similarity (Observation~2)}
\label{app:state}

For each token position $s$, the per-token cosine similarity at layer $\ell$ between consecutive steps is
\begin{equation*}
c_\ell^t[s] \;=\; \frac{\bigl\langle H_\ell^t[s,:],\, H_\ell^{t+1}[s,:]\bigr\rangle}{\bigl\|H_\ell^t[s,:]\bigr\|_2 \,\cdot\, \bigl\|H_\ell^{t+1}[s,:]\bigr\|_2} \in [-1, 1].
\end{equation*}
The reported {$\geq 0.97$} value is the grand mean
\begin{equation*}
\bar c \;=\; \frac{1}{L\,S\,(T-1)} \sum_{\ell=1}^{L} \sum_{s=1}^{S} \sum_{t=0}^{T-2} c_\ell^t[s],
\end{equation*}
computed in float32 to avoid BF16 dot-product rounding error.

\subsection{Expert Sparsity (Observation~3)}
\label{app:sparsity}

For each expert $e \in \{1,\ldots,E\}$, the per-expert token-assignment count at layer $\ell$ and step $t$ is
\begin{equation*}
n_\ell^t[e] \;=\; \bigl|\bigl\{(s, j) : R_\ell^t[s, j] = e\bigr\}\bigr|.
\end{equation*}
The \emph{top-10 token-mass share} $\rho_\ell^t$ is the sum of the ten largest entries of $n_\ell^t[\,\cdot\,]$ divided by the total mass $\sum_{e=1}^{E} n_\ell^t[e]$; the \emph{idle expert count} is $m_\ell^t = \bigl|\{e : n_\ell^t[e] = 0\}\bigr|$. The reported values are averages of $\rho_\ell^t$ and $(E-m_\ell^t)$ over all layers and denoising steps at the benchmark operating point (32 steps, generation length 384): top-10 mass share $\sim$65\%, active set $\sim$150 of 256 experts. The active-set size is strongly sequence-length dependent: at the short profiling configuration (generation length 64) only $\sim$78 experts per layer are active, so working-set figures must be read against the generation length at which they were measured. The 12~GB per-layer cache capacity ($\sim$78 experts) is a distinct, memory-derived quantity and should not be confused with the active-set size.

%==========================================================================
\section{Prefetch Algorithm}
\label{app:algorithm}

Algorithm~\ref{alg:prefetch} gives the full pseudocode for the per-step prefetch--compute pipeline described in prose in Section~\ref{sec:method:prefetch}, covering the initial lookahead staging, the per-layer compute sweep with stream synchronization, and buffer recycling for layers $\ell{+}D$ onward.

\begin{figure*}[t]
  \centering
  \includegraphics[width=0.7\textwidth]{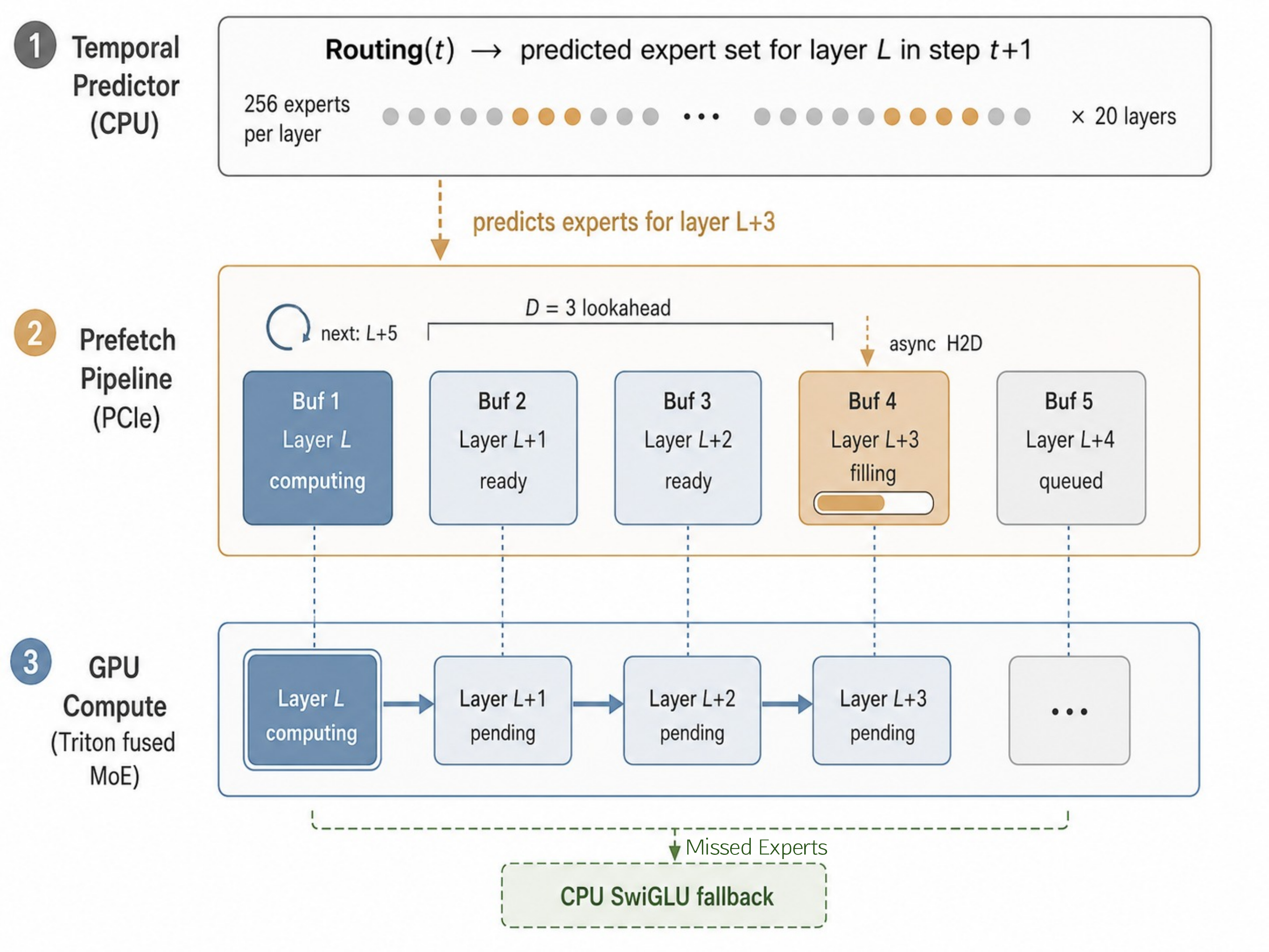}
  \caption{Temporal expert weight prefetching. The CPU-side predictor uses the previous step's per-layer routing $E_\ell^t$ to forecast $\hat E_\ell^{t+1}$ and issues asynchronous host-to-device copies on independent transfer streams into a pool of on-device buffers, each staged $D$ layers ahead of the GPU compute stream. A buffer cycles back into the pool immediately after its layer completes, allowing PCIe transfers to overlap with compute in steady state.}
  \label{fig:prefetch}
\end{figure*}

\begin{algorithm*}[t]
\small
\caption{Per-step prefetch--compute pipeline (denoising step $t{+}1$).}
\label{alg:prefetch}
\begin{algorithmic}[1]
\Require hidden states $H_{t+1}$; MoE layers $\ell_1{<}\cdots{<}\ell_L$; buffers $\mathcal{B}_1,\ldots,\mathcal{B}_N$; streams $\sigma_c,\sigma_t^1,\ldots,\sigma_t^N$; previous routing $E_\ell^t$; lookahead $D$.
\For{$i = 1$ to $\min(D,N)$} \Comment{stage initial lookahead}
  \State $\ell \gets \ell_i$; assign $\mathcal{B}_i$ to $\ell$; $P_\ell \gets E_\ell^t$
  \For{$e \in P_\ell$}
    \State \texttt{cudaMemcpyAsync}(cpu$[\ell,e] \to$ slot in $\mathcal{B}_i$) on $\sigma_t^i$
  \EndFor
  \State record event \texttt{evt}$[\ell]$ on $\sigma_t^i$
\EndFor
\For{$j = 1$ to $L$} \Comment{compute sweep}
  \State $\ell \gets \ell_j$; $\mathcal{B} \gets$ buffer assigned to $\ell$
  \State attention, gating on $\sigma_c \Rightarrow$ \texttt{topk\_idx}; $E_\ell^{t+1} \gets \mathrm{unique}(\texttt{topk\_idx})$
  \State $\sigma_c.\texttt{wait}(\texttt{evt}[\ell])$ \Comment{GPU-side wait}
  \For{$e \in E_\ell^{t+1}\setminus\mathrm{loaded}(\mathcal{B})$} \Comment{residual miss}
    \State sync-load $e$ into $\mathcal{B}$, \textbf{or} CPU SwiGLU($e$)
  \EndFor
  \State sort tokens by expert; build offsets; dispatch fused MoE on $\sigma_c$
  \State release $\mathcal{B}$; assign to $\ell_{j+D}$; start prefetch on $\sigma_t^{\mathrm{idx}(\mathcal{B})}$
  \State $E_\ell^t \gets E_\ell^{t+1}$ \Comment{for next step's prediction}
\EndFor
\end{algorithmic}
\end{algorithm*}

%==========================================================================
\section{Preliminary Ablations}
\label{app:ablations}

This appendix reports the ablation data summarized in Section~\ref{sec:setup:ablations}. Each subsection establishes the operating point for one hyperparameter; the corresponding interpretation appears in Section~\ref{sec:analysis}. 

\subsection{Output Cache: $\tau$ Sweep}
\label{app:abl:tau}

We compared the output cache at three cosine-similarity thresholds, with and without the Jacobian Taylor correction, on a tuning subset of MMLU and GSM8K. We tune on this subset rather than the coding benchmarks to fix $\tau$ on data disjoint from the headline coding/reasoning evaluation; the resulting operating point ($\tau{=}0.99$) transfers without modification (Section~\ref{sec:results}). Table~\ref{tab:abl:tau} reports speedup and downstream accuracy; Table~\ref{tab:abl:tau-cat} reports the resulting token-category distribution.

\begin{table}[h]
\centering
\small
\caption{Output cache: speedup and accuracy at varying cosine-similarity thresholds. \emph{Conditions differ from the main evaluation:} this sweep was run during preliminary development on LLaDA2.1-mini with the full model GPU-resident (no memory cap, so no PCIe bottleneck), on a tuning subset of 30 MMLU and 30 GSM8K questions at generation length 256. Speedups are therefore compute-only and accuracies are small-sample; neither is comparable to the full-set, 12~GB results of Section~\ref{sec:results}. The sweep is reported only to justify the choice $\tau{=}0.99$, which transfers to the constrained regime without modification.}
\label{tab:abl:tau}
\begin{tabular}{lccc}
\toprule
\textbf{Config} & \textbf{Speedup} & \textbf{MMLU} & \textbf{GSM8K} \\
\midrule
Baseline & 1.00$\times$ & 100.0\% & 96.7\% \\
naive $\tau{=}0.99$ & \textbf{2.3--2.8$\times$} & 100.0\% & 96.7\% \\
naive $\tau{=}0.995$ & 1.62$\times$ & 100.0\% & 96.7\% \\
naive $\tau{=}0.999$ & 1.34$\times$ & 100.0\% & 86.7\% \\
Jacobian $\tau{=}0.99$ & 0.49$\times$ & 100.0\% & 86.7\% \\
\bottomrule
\end{tabular}
\end{table}

\begin{table}[h]
\centering
\small
\setlength{\tabcolsep}{3pt}
\caption{Token-category distribution by $\tau$ (naive reuse), measured under preliminary conditions (full residency, 50 prompts, generation length 128). Tightening $\tau$ from 0.99 to 0.999 shifts $\sim$6~pp from Cat~A reuse to Cat~C full recomputation without measurable accuracy benefit. \emph{Hit}: fraction of expert evaluations served from cache; \emph{Skip}: fraction of per-expert weight loads avoided.}
\label{tab:abl:tau-cat}
\begin{tabular}{lcccccc}
\toprule
$\tau$ & \textbf{A} & \textbf{B} & \textbf{C} & \textbf{Cold} & \textbf{Hit} & \textbf{Skip} \\
\midrule
0.99  & 72.4\% & 2.4\% & 16.4\% & 8.7\% & 73.7\% & 74.4\% \\
0.995 & 70.5\% & 1.3\% & 19.5\% & 8.7\% & 69.8\% & 71.6\% \\
0.999 & 66.8\% & 0.2\% & 24.4\% & 8.6\% & 64.5\% & 67.0\% \\
\bottomrule
\end{tabular}
\end{table}

\noindent\textbf{Operating point.} We adopt naive reuse at $\tau{=}0.99$. Tighter thresholds force tokens into full recomputation without quality benefit; the Jacobian correction (three additional matmuls per Cat~B token) costs more than it saves.

\subsection{Prefetch Lookahead Depth}
\label{app:abl:lookahead}

We swept the layer-level lookahead $D \in \{3, 6, 12\}$ across both models and three GPU memory budgets on GSM8K (300--400 problems per cell, generation length 384, L40S). The 16 and 24~GB budgets are included to confirm that $D{=}3$ is optimal not only under the 12~GB target but as memory is relaxed---i.e., the operating point is not a 12~GB artifact. Results in Table~\ref{tab:abl:lookahead}.

\begin{table}[h]
\centering
\small
\setlength{\tabcolsep}{5pt}
\caption{Lookahead depth ablation on GSM8K (p50 ms/step). $D{=}3$ is optimal at every (model, GPU-budget) configuration, including as the budget is relaxed beyond the 12~GB target.}
\label{tab:abl:lookahead}
\begin{tabular}{llccc}
\toprule
\textbf{Model} & \textbf{GPU} & \textbf{$D{=}3$} & \textbf{$D{=}6$} & \textbf{$D{=}12$} \\
\midrule
LLaDA2.0-mini & 12\,GB & \textbf{790} & 918 & 1074 \\
              & 16\,GB & \textbf{783} & 835 & 1034 \\
              & 24\,GB & \textbf{791} & 820 & 955 \\
\midrule
LLaDA2.1-mini & 12\,GB & \textbf{735} & 865 & 992 \\
              & 16\,GB & \textbf{744} & 770 & 932 \\
              & 24\,GB & \textbf{735} & 776 & 875 \\
\bottomrule
\end{tabular}
\end{table}

\noindent\textbf{Operating point.} We adopt $D{=}3$ with a 5-buffer pool. Increasing $D$ from 3 to 12 adds 19--36\% latency while changing hit rate by at most 1--2~pp; the reason is analyzed in Section~\ref{sec:analysis}.

\subsection{Prediction Strategy}
\label{app:abl:predictor}

% REVIEW NOTE: the predictor-ablation result files on disk (dllms/results/pred_temporal_*_24gb.json) are 3 prompts, generation length 128, 24 GB; confirm the 12 GB stated here.
We compared the pure-temporal predictor (previous-step routing) against three augmented variants (LLaDA2.1-mini, L40S, 12~GB). This setup isolates prediction quality from prefetch-pool sizing, which the lookahead sweep (Appendix~\ref{app:abl:lookahead}) isolates separately. Results in Table~\ref{tab:abl:predictor}.

\begin{table}[h]
\centering
\footnotesize
\setlength{\tabcolsep}{3pt}
\caption{Prediction strategy ablation. Pure temporal sits on the Pareto frontier: every augmented predictor raises hit rate by 1--4~pp but slows ms/step. \emph{Cap}: maximum number of experts prefetched per layer per step.}
\label{tab:abl:predictor}
\begin{tabular}{lcccc}
\toprule
\textbf{Policy} & \textbf{Cap} & \textbf{ms/step} & \textbf{$\Delta$~ms} & \textbf{Hit~\%} \\
\midrule
Temporal (ours)         & 160 & \textbf{672--693} & ref      & 89--91\% \\
Temporal-deep ($D{=}2$) & 180 & 690--752          & $+$1--8\%  & 90--94\% \\
EMA workload ($K{=}8$)  & 180 & 693--738          & $+$3--6\%  & 93--94\% \\
Temporal-spatial        & 200 & 1100--1155        & $+$64\%    & 91--95\% \\
\bottomrule
\end{tabular}
\end{table}

\noindent\textbf{Operating point.} We adopt the pure-temporal predictor. The augmented predictors enlarge the prefetched set and therefore the PCIe volume per step, which dominates the small hit-rate gains.

%==========================================================================
\section{Benchmark Details}
\label{app:benchmarks}

This appendix expands on the benchmark summaries in Section~\ref{sec:setup:benchmarks}, giving the full per-benchmark prompting protocol, generation length where applicable, and scoring rule.

\paragraph{HumanEval+~\cite{liu2023humanevalplus}.}
164 Python function-completion problems with $\sim$80$\times$ more unit tests per problem than the original HumanEval, making the benchmark substantially more robust to surface-level pattern matching. Generation length 512 tokens, average $\sim$215 tokens emitted per problem. Zero-shot instruction-tuned chat prompting; scored by unit-test pass rate (pass@1).

\paragraph{LiveCodeBench (easy+medium)~\cite{jain2024lcb}.}
310 contamination-resistant competitive-programming problems drawn from LeetCode, AtCoder, and Codeforces. Generation length 1024 tokens, average $\sim$310 tokens emitted. Zero-shot instruction-tuned chat prompting; scored by unit-test pass rate (pass@1).

\paragraph{ARC-C~\cite{clark2018arc}.}
1{,}172 grade-school science multiple-choice questions with four options each, selected to resist surface-pattern shortcuts and to require commonsense plus scientific knowledge. Zero-shot instruction prompting $\to$ letter answer.

\paragraph{MMLU~\cite{hendrycks2021mmlu}.}
Multiple-choice questions across 57 subjects spanning STEM, humanities, social sciences, and applied/professional domains, designed to assess breadth of academic and professional knowledge. 4-shot chain-of-thought prompting $\to$ letter A/B/C/D.

\paragraph{TriviaQA~\cite{joshi2017triviaqa}.}
1{,}000 open-domain factoid questions drawn from trivia leagues and quiz bowls. 4-shot string prompting $\to$ free-form short answer; scored by exact-match and token-F1 against an alias set per question.

\paragraph{GSM8K~\cite{cobbe2021gsm8k}.}
1{,}319 grade-school math word problems requiring multi-step arithmetic and chain-of-thought reasoning. Generation length 384 tokens, 4-shot CoT prompting; scored by exact-match on the final numerical answer.

\paragraph{BBH~\cite{suzgun2022bbh}.}
6{,}511 reasoning examples across 23 subtasks (yes/no, multiple-choice letter, free-form text, integers, lists) drawn from BIG-Bench and selected to target compositional and multi-step reasoning. Zero-shot chain-of-thought prompting; per-subtask scoring rules follow the original release.

%==========================================================================
\section{Cache-Category Breakdown}
\label{app:cat-table}

\begin{figure*}[t]
\centering
\includegraphics[width=0.95\textwidth]{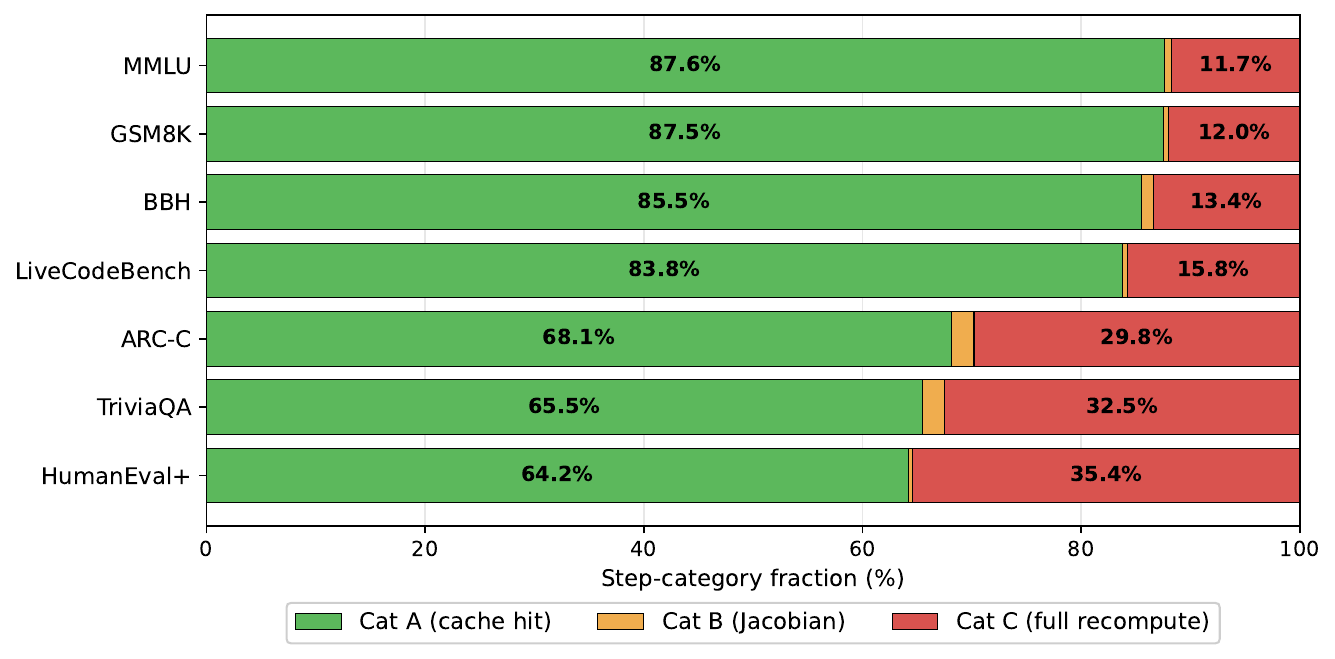}
\caption{Cache-category breakdown under combined caching, averaged across (model, GPU) cells for each benchmark. Cat~A: full output-cache reuse; Cat~B: partial reuse; Cat~C: full recompute. \emph{Chain-of-thought and long-form benchmarks (LCB, BBH, GSM8K, MMLU) reach 84--88\% cache hits per step; benchmarks with shorter or more discrete outputs (HumanEval+, TriviaQA, ARC-C) reach 64--68\%.}}
\label{fig:cat}
\end{figure*}

Table~\ref{tab:cat} reports the exact per-benchmark Cat~A/B/C fractions averaged over GPU cells; the same numbers are visualized in Figure~\ref{fig:cat}. The category fractions are essentially GPU-invariant (matches within 0.5~pp between L40S and A6000).

\begin{table}[h]
\centering
\small
\setlength{\tabcolsep}{8pt}
\caption{Cache-category breakdown under combined caching, averaged across (model, GPU) cells. Cat~A: full output-cache reuse; Cat~B: partial reuse; Cat~C: full recompute.}
\label{tab:cat}
\begin{tabular}{lccc}
\toprule
\textbf{Benchmark} & \textbf{Cat~A} & \textbf{Cat~B} & \textbf{Cat~C} \\
\midrule
HumanEval+    & 64.2\%     & 0.4\%  & 35.4\%      \\
LiveCodeBench & 83.8\%     & 0.4\%  & 15.8\%      \\
GSM8K         & 87.5\%     & 0.5\%  & 12.0\%      \\
ARC-C         & 68.1\%     & 2.1\%  & 29.8\%      \\
MMLU          & 87.6\%     & 0.7\%  & 11.7\%      \\
TriviaQA      & 65.5\%     & 2.0\%  & 32.5\%      \\
BBH           & 85.5\%     & 1.1\%  & 13.4\%      \\
\bottomrule
\end{tabular}
\end{table}

%==========================================================================
\section{Quality Preservation}
\label{app:accuracy}

Table~\ref{tab:accuracy} reports the per-cell quality metric for each measured (benchmark, model, GPU) cell, comparing the accelerate baseline against combined caching. The metric is pass@1 for the code benchmarks (HumanEval+, LCB), letter-match accuracy for the MCQ benchmarks (ARC-C, MMLU), and exact-match accuracy on the parsed answer for the remaining free-form benchmarks (TriviaQA, GSM8K, BBH); the exact prompting protocols and scoring rules are in Section~\ref{sec:setup:benchmarks} and Appendix~\ref{app:benchmarks}. Grouping by benchmark family, the largest combined--accel gap is $\leq$3.5~pp on code, $\leq$2.8~pp on free-form recall (TriviaQA), $\leq$2.2~pp on MCQ and compositional reasoning (ARC-C, MMLU, BBH), and $\leq$1.5~pp on math (GSM8K). Every gap is within the run-to-run variance quantified in Section~\ref{sec:results:speedup}; on the one cell measured at full size with a confidence interval (GSM8K, A6000, LLaDA2.0-mini, $n{=}1{,}319$), combined caching lands inside the baseline interval. No cell shows a measurable quality loss.

\begin{table}[h]
\centering
\footnotesize
\setlength{\tabcolsep}{3pt}
\caption{Per-cell output quality at the 12~GB budget. pass@1 for HumanEval+ ($n{=}164$) and LCB ($n{=}310$); letter-match accuracy for ARC-C ($n{=}1{,}172$) and MMLU ($n{=}14{,}042$); exact-match accuracy for TriviaQA ($n{=}1{,}000$), GSM8K ($n{=}1{,}319$), and BBH ($n{=}6{,}511$). The GSM8K A6000/2.0-mini accel cell is the full 1{,}319-problem run: 89.69\%, 95\% CI [88.05, 91.33], against 90.45\% for combined caching. The GSM8K L40S Accelerate baselines remain sub-sampled ($n{=}300$ each) because the Accelerate baseline runs prohibitively slowly at this budget.}
\label{tab:accuracy}
\begin{tabular}{llcccc}
\toprule
 & & \multicolumn{2}{c}{\textbf{LLaDA2.0-mini}} & \multicolumn{2}{c}{\textbf{LLaDA2.1-mini}} \\
\cmidrule(lr){3-4}\cmidrule(lr){5-6}
\textbf{Benchmark} & \textbf{GPU} & accel & combined & accel & combined \\
\midrule
HumanEval+ & L40S  & 82.9 & 80.5 & 81.7 & 82.9 \\
           & A6000 & 79.9 & 80.5 & 82.9 & 79.9 \\
\midrule
LCB        & L40S  & 49.4 & 49.7 & 39.4 & 42.9 \\
           & A6000 & 47.1 & 48.1 & 41.3 & 41.9 \\
\midrule
GSM8K      & L40S  & 90.3 & 90.0 & 84.3 & 85.6 \\
           & A6000 & 89.7 & 90.45 & 83.3 & 84.5 \\
\midrule
ARC-C      & L40S  & 89.0 & 88.5 & 87.6 & 87.8 \\
           & A6000 & 88.9 & 88.9 & 87.7 & 87.2 \\
\midrule
TriviaQA   & L40S  & 44.3 & 44.4 & 43.2 & 44.6 \\
           & A6000 & 46.2 & 43.4 & 44.2 & 44.8 \\
\midrule
MMLU       & L40S  & 74.4 & 76.6 & 58.0 & 58.2 \\
           & A6000 & 70.3 & 68.9 & 54.2 & 52.7 \\
\midrule
BBH        & L40S  & 50.7 & 49.2 & 48.8 & 48.6 \\
           & A6000 & 49.5 & 50.3 & 49.1 & 49.2 \\
\bottomrule
\end{tabular}
\end{table}

%==========================================================================
\section{Potential Risks}
\label{app:risks}
Our system accelerates inference of existing publicly released MoE-DLMs without altering their parameters or routing rule. Combined caching preserves task accuracy within measured run-to-run variance (Appendix~\ref{app:accuracy}), although approximate output reuse can change individual generations. The principal consideration is dual-use: lowering the memory floor for capable MoE-DLMs widens access for research and education but also marginally lowers the barrier to misuse. The increment is small relative to existing access via larger GPUs or hosted APIs.

%==========================================================================
\section{Ethics Statement}
\label{app:ethics}
We do not train or fine-tune any model, collect or annotate any data, or involve human subjects. Experiments use publicly released benchmarks and pre-trained LLaDA weights via HuggingFace. AI assistants were used for writing and code review; all technical content was authored and verified by the listed contributors.

%==========================================================================
\section{Artifact Licenses and Intended Use}
\label{app:licenses}
HumanEval+, GSM8K, MMLU, and BBH are released under MIT; LiveCodeBench under CC~BY~4.0; ARC-C under CC~BY-SA~4.0; TriviaQA under Apache~2.0. LLaDA2.0-mini and 2.1-mini are distributed through HuggingFace under their model-card licenses. Our implementation builds on HuggingFace Transformers, Accelerate, PyTorch, and Triton, with the HumanEval+ harness derived from \texttt{evalplus} (Apache~2.0). All artifacts are used for inference benchmarking only, consistent with each artifact's intended use.

\end{document}